\documentclass[runningheads]{llncs}

\usepackage{graphicx}
\usepackage{booktabs}
\usepackage{hyperref}
\usepackage{xcolor}
\usepackage{amsmath}
\usepackage{amssymb}
\usepackage{algorithm}
\usepackage{algpseudocode}
\usepackage{tikz}
\usetikzlibrary{shapes, arrows.meta, positioning, calc, fit, backgrounds, shapes.geometric, shadows}

\usepackage{filecontents}

\title{VALOR: Value-Aware Revenue Uplift Modeling with Treatment-Gated Representation for B2B Sales}
\titlerunning{VALOR: Value-Aware Revenue Uplift Modeling}

\author{Vamshi Guduguntla \and Kavin Soni \and Debanshu Das}
\authorrunning{V. Guduguntla et al.}
\institute{Google, California, USA\\
\email{\{vamshigud, kavinsoni, debanshu\}@google.com}}

\begin{document}

\maketitle

\begin{abstract}
B2B sales organizations must identify "persuadable" accounts within zero-inflated revenue distributions to optimize expensive human resource allocation. Standard uplift frameworks struggle with treatment signal collapse in high-dimensional spaces and a misalignment between regression calibration and the ranking of high-value "whales." We introduce VALOR (Value Aware Learning of Optimized (B2B) Revenue), a unified framework featuring a Treatment-Gated Sparse-Revenue Network that uses bilinear interaction to prevent causal signal collapse. The framework is optimized via a novel Cost-Sensitive Focal-ZILN objective that combines a focal mechanism for distributional robustness with a value-weighted ranking loss that scales penalties based on financial magnitude.

To provide interpretability for high-touch sales programs, we further derive Robust ZILN-GBDT, a tree based variant utilizing a custom splitting criterion for uplift heterogeneity. Extensive evaluations confirm VALOR's dominance, achieving a 20\% improvement in rankability over state-of-the-art methods on public benchmarks and delivering a validated 2.7$\times$ increase in incremental revenue per account in a rigorous 4-month production A/B test.
\end{abstract}

\section{Introduction}
In the highly competitive domain of enterprise cloud computing, sales organizations face a critical resource allocation dilemma \cite{reutterer2006dynamic}. While the potential customer base (the ``Long Tail'') is vast, the capacity for high-touch sales intervention is strictly finite  \cite{zhang2021unified,gutierrez2017causal,goldenberg2020free,albert2021commerce}. Unlike Business-to-Consumer (B2C) e-commerce, where algorithmic bidding systems can target millions of users instantaneously at negligible marginal cost \cite{zhang2021unified,devriendt2020learning}, Business-to-Business (B2B) sales cycles are driven by expensive human capital—Business Development Representatives (BDRs). A single intervention often involves weeks of relationship building,technical validation, and negotiation. Consequently, the cost of a ``False Positive''—allocating a BDR to a client who would have purchased organically (``Sure Things'') or one who will never purchase (``Lost Causes'')—is not merely a lost impression, but a significant waste of operational budget \cite{efin2023,goldenberg2020free,albert2021commerce,zhou2023direct}.

This operational reality requires a paradigm shift from traditional propensity modeling ($P(Y|X)$) to causal uplift modeling ($P(Y|T=1) - P(Y|T=0)$) \cite{reutterer2006dynamic,zhang2021unified,gutierrez2017causal,curth2021nonparametric}. Although propensity models effectively identify high-value accounts \cite{kane2014mining}, they fail to disentangle the effect of treatment from the likelihood of purchase at the time of assignment \cite{yao2018representation,louizos2017causal,curth2021nonparametric,athey2015machine}. In practice, this results in sales teams ``skimming the cream''—targeting high-velocity accounts that generate revenue regardless of intervention—thereby inflating perceived ROI while delivering negligible incremental value.

\subsection{The Gap in Current Methodologies}
Despite the promise of uplift modeling, its application to B2B revenue optimization faces two fundamental challenges. \textbf{Counterfactual Gradient Collapse:} B2B revenue features a dominant zero-mass ($>80\%$ non-conversion) and a heavy-tailed positive distribution \cite{rerum2024}. Standard meta-learners optimizing Mean Squared Error (MSE) \cite{kunzel2019metalearners,nie2021quasi} suffer a \textit{collapse toward the global mean}, predicting near-zero uplift globally and failing to isolate rare ``Persuadables.'' Furthermore, statistical calibration diverges from decision-making utility \cite{dfcl2023}; existing symmetric losses fail to penalize the severe operational cost of mis-ranking high-value ``whales'' below median spenders. \textbf{Prognostic Dominance:} In high-dimensional enterprise telemetry, simply concatenating the treatment indicator $T$ with features $X$ leads to pervasive underutilization of the causal signal \cite{efin2023,sun2023robustness}. Strong prognostic main effects dominate the latent representation \cite{yao2018representation,louizos2017causal}, regularizing the heterogeneous treatment effect (HTE) toward zero—a phenomenon termed \textit{regularization bias} \cite{shalit2017estimating}. This basic concatenation lacks the expressivity for the bilinear interactions \cite{calder2003feature,efin2023,rendle2011fast,juan2016field,sun2023robustness} essential for modeling conditional B2B usage triggers, degrading the model into a mere propensity estimator. 

To bridge this gap, a framework must move beyond simple calibration toward a paradigm that is simultaneously \textit{distribution-aware} and \textit{interaction-explicit}, motivating the design of VALOR.

\subsection{Our Contributions}
To address prognostic dominance and counterfactual gradient collapse in capacity-constrained B2B revenue optimization, we propose VALOR. Our key contributions are:

\begin{itemize}
    \item \textbf{Treatment-Gated Network \& Cost-Sensitive Objective:} We introduce a deep learning framework coupling a bilinear gating mechanism (to mitigate the ``vanishing treatment'' signal) with a novel Focal-ZILN and Value-Weighted Ranking loss. This aligns model optimization directly with the asymmetric financial risk of mis-ranking high-value enterprise whales.
    
    \item \textbf{Interpretable ZILN-GBDT:} To satisfy the strict explainability requirements of human-driven sales programs, we derive a tree-based variant utilizing a custom uplift heterogeneity splitting criterion and adaptive Bayesian smoothing for sparse leaf nodes.
    
    \item \textbf{End-to-End Causal System Architecture:} We detail the production deployment of VALOR, featuring drift-aware MLOps, a capacity-aware hybrid routing strategy, and a rigorous closed-loop outcome tracking pipeline that resolves the 90-day B2B sales lifecycle into clean causal feedback.
    
    \item \textbf{Proven Industrial Impact:} Offline, VALOR achieves a 20\% improvement in rankability (Qini) over state-of-the-art baselines. Online, a 4-month Randomized Controlled Trial (RCT) on a large cloud platform validates a 2.7$\times$ increase in incremental revenue per account (\$1185 vs \$445), projecting an estimated \$30M annualized lift.
\end{itemize}

\section{Related Work}
Traditional uplift modeling has evolved from meta-learners \cite{kunzel2019metalearners} to deep representation architectures (e.g., TARNet, CFRNet \cite{shalit2017estimating}) that utilize Integral Probability Metrics to mitigate selection bias \cite{wu2023stable}. However, in high-dimensional settings, standard backbones suffer from ``regularization bias'' where strong prognostic features drown out subtle treatment effects \cite{efin2023}. While frameworks like EFIN \cite{efin2023,guo2017deepfm,wang2021dcn} and UMLC \cite{sun2025robust} address this via explicit feature interactions and context grouping for B2C scenarios, they rely on dense patterns distinct from sparse B2B telemetry. Furthermore, modeling continuous revenue introduces the challenge of zero-inflation. RERUM \cite{rerum2024} adapted the Zero-Inflated LogNormal (ZILN) loss and ranking objectives for e-commerce, but standard ZILN optimization frequently collapses under the extreme sparsity ($>80\%$ zeros) characteristic of enterprise sales \cite{gubela2020response,DBLP:conf/amcis/GubelaL20}. Finally, Decision-Focused Learning (DFL) demonstrates that minimizing standard prediction errors (MSE) does not guarantee optimal downstream decisions \cite{dfcl2023}. While methods like DFCL \cite{dfcl2023} use Lagrangian duality for budget optimization, VALOR adopts DFL principles within a ``predict-then-rank'' paradigm by introducing a value-weighted ranking loss that acts as a differentiable proxy for financial ROI, effectively bridging the gap between regression accuracy and operational yield.


\section{Problem Formulation}
We formulate B2B sales resource allocation using the standard Neyman-Rubin potential outcomes framework \cite{rubin2005causal}, aiming to estimate the Conditional Average Treatment Effect (CATE): $\hat{\tau}(x) = \mathbb{E}[Y | X=x, T=1] - \mathbb{E}[Y | X=x, T=0]$.

\subsection{The Zero-Inflated Distribution}
In the targeted SMB segment, revenue $Y$ is semi-continuous, characterized by frequent zeros and heavy-tailed positive values. We model this via a mixture distribution $p(y | \theta) = (1 - \rho) \delta_0(y) + \rho \cdot \text{LogNormal}(y; \mu, \sigma^2)$.
Here, $\rho$ is the conversion probability, $\mu$ and $\sigma$ parameterize the positive log-normal component, and $\delta_0$ is the Dirac delta at zero. The objective is to estimate $\hat{\tau}(x)$ such that ranking accounts by this score maximizes the Area Under the Uplift Curve (AUUC).

\section{Proposed Architecture: VALOR}
To address the limitations of the incumbent model and the specific deficiencies of standard ZILN losses in highly imbalanced B2B settings, we introduce VALOR. 
\begin{figure*}[t]
  \centering
  \includegraphics[width=\textwidth]{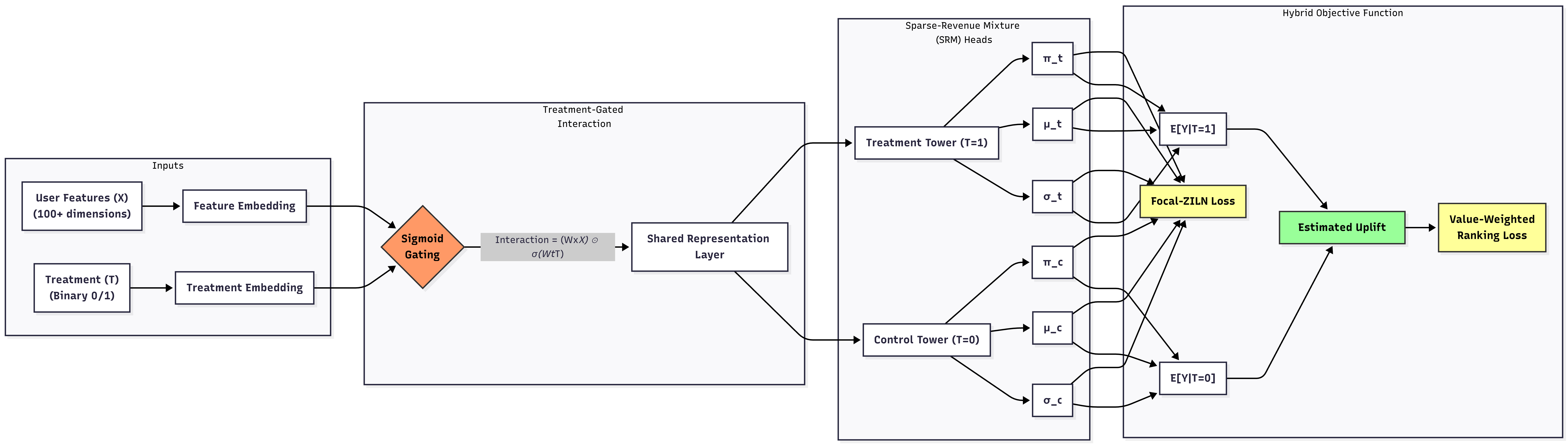}
\caption{The VALOR training pipeline processes features ($X$) and treatment ($T$) through three stages: (1) A Treatment-Gated Interaction Module uses a sigmoid gate to dynamically re-weight features, mitigating the ``vanishing treatment'' signal; (2) Sparse-Revenue Mixture Heads decouple the decision ($\pi$) and value ($\mu, \sigma$) processes to model zero-inflated revenue; and (3) A Hybrid Objective combines Focal-ZILN and Value-Weighted Ranking losses to ensure distributional robustness and maximize top-decile revenue capture.}
  \label{fig:valor_architecture}
\end{figure*}


\subsection{The Treatment-Gated Sparse-Revenue Network}
VALOR addresses the B2B Sparse-Revenue Mixture (SRM) by formally disentangling the \textit{decision process} (conversion) from the \textit{value process} (magnitude). A \textbf{Treatment-Gated Backbone} dynamically re-weights features $\Phi(X, T)$ based on intervention, feeding bifurcated heads that output three SRM components: (1) \textbf{Hurdle Probability ($\pi$):} The probability of overcoming the ``zero mass'' ($P(Y>0)$); (2) \textbf{Conditional Value ($\mu$):} The expected log-scale revenue magnitude; and (3) \textbf{Value Volatility ($\sigma$):} The aleatoric uncertainty, modeled to penalize noise-driven estimates.
The expected value for branch $k \in \{0, 1\}$ is reconstructed as $\mathbb{E}[y_k] = \pi_k \cdot \exp\left(\mu_k + \frac{\sigma_k^2}{2}\right)$.

\subsection{Proposed Hybrid Objective: Cost-Sensitive Focal-ZILN}
Standard ZILN losses \cite{rerum2024}\cite{wang2019deep} struggle in sparse B2B domains due to (1) \textit{gradient domination} by abundant zeros, which biases models toward non-conversion, and (2) \textit{cost-insensitive ranking}, where mis-ranking high-value "whales" is treated identically to low-value errors. To address this, we propose a hybrid objective combining a focal mechanism for sparsity and value-weighting for ROI alignment.

\subsubsection{Focal-ZILN Loss ($\mathcal{L}_{FL-ZILN}$)}
To prevent "easy negatives" (non-spenders) from overwhelming the gradient, we adapt Focal Loss for the propensity component $\mathcal{L}_{prop}$. With focusing parameter $\gamma \ge 0$ and balance factor $\alpha$:

\begin{equation}
    \mathcal{L}_{prop} = \begin{cases} 
    -\alpha (1-p_i)^\gamma \log(p_i) & \text{if } y_i > 0 \\
    -(1-\alpha) p_i^\gamma \log(1-p_i) & \text{if } y_i = 0 
    \end{cases}
\end{equation}

The regression component $\mathcal{L}_{rev}$ remains the log-likelihood for positive instances ($y_i > 0$), and the total distribution loss is $\mathcal{L}_{FL-ZILN} = \mathcal{L}_{prop} + \mathcal{L}_{rev}$:

\begin{equation}
    \mathcal{L}_{rev} = \mathbb{I}(y_i > 0) \cdot \left( \frac{1}{2} \left( \frac{\log y_i - \mu_i}{\sigma_i} \right)^2 + \log(\sigma_i \sqrt{2\pi}) \right)
\end{equation}

\subsubsection{Value-Weighted Ranking Loss ($\mathcal{L}_{V-Rank}$)}
To prioritize high-value accounts, we introduce a cost-sensitive pairwise loss. Unlike standard ranking objectives, we scale penalties by the financial magnitude of the error. Let $z_i$ be the transformed outcome; we define a dynamic pair weight $w_{ij} = \log(1 + |z_i - z_j|)$. This ensures that catastrophic inversions dominate the optimization. The final ranking loss is $\mathcal{L}_{V-Rank} = \sum_{i,j} w_{ij} \cdot \log \left( 1 + \exp \left( - \text{sign}(z_i - z_j)(\hat{\tau}_i - \hat{\tau}_j) \right) \right)$.

\subsection{Treatment-Gated Representation Learning}
In high-dimensional B2B feature spaces, the treatment signal is often drowned out by the strong main effects of prognostic features. Standard architecture designs that simply concatenate treatment and features ($[X, T]$) fail here; deep networks tend to latch onto the dominant prognostic signal while treating the sparse treatment indicator as negligible noise—a phenomenon known as the ``vanishing treatment'' problem \cite{efin2023}\cite{zhong2022descn}. To address this, we introduce a Treatment-Gated Interaction Module. Rather than simple concatenation, we use the treatment embedding to dynamically re-weight the feature representation via a bilinear gating operation: $\text{Interaction}(X, T) = (W_x X + b_x) \odot \sigma(W_t T + b_t)$, where $\odot$ denotes the element-wise Hadamard product and $\sigma$ is a sigmoid activation.

\textbf{Why Gating Outperforms Concatenation:} This mechanism functions as a differentiable, treatment-conditional feature selector. Unlike the additive interaction approach in EFIN \cite{efin2023}, which relies on concatenating embeddings and subsequent layers to implicitly learn interactions, our multiplicative gate enables the treatment embedding $\Psi(T)$ to explicitly ``zero-out'' irrelevant feature subspaces (driving $\sigma \to 0$) before they enter the regressor. For example, if a treatment targets only ``Compute'' usage, the gate suppresses high-variance ``Storage'' features that act as noise. This suppression of irrelevant prognostic variance is critical for the \textbf{Focal-ZILN} loss to converge, ensuring gradient updates focus exclusively on feature subspaces driving heterogeneous lift.

\subsection{Interpretable Variant: Robust ZILN-GBDT}
While VALOR's deep network maximizes revenue for automated high-volume routing, premium account management by human agents demands strict \textit{interpretability}. We therefore distill VALOR's core principles---zero-inflation awareness and uplift maximization---into a lightweight variant: Robust ZILN-GBDT. Unlike standard trees that maximize homogeneity, our formulation maximizes treatment effect heterogeneity via a custom splitting criterion based on the Euclidean distance of ZILN uplift between child nodes. We pair this with Adaptive Bayesian Smoothing to prevent variance collapse in sparse leaves (full mathematical derivations and algorithms are provided in the Supplementary Material). This yields a bifurcated deployment strategy: VALOR (DNN) for scalable automation, and ZILN-GBDT (Tree) for interpretable human decision support.


\section{Experimental Evaluation}
We execute a multi-faceted evaluation strategy using public benchmarks and proprietary data to validate the proposed VALOR and the proposed ZILN-GBDT algorithm. 

\subsection{Datasets}
We evaluate VALOR on two datasets: a high-fidelity synthetic benchmark and a proprietary real-world log. \textbf{Synthetic Dataset (B2B-Mimic):} Since counterfactuals are unobservable, we adapted the UMLC generation protocol \cite{sun2025robust}. To mirror B2B accounts engaging with multiple product lines (e.g., Compute, AI), we generated 236,421 accounts with 100 features (34 binary, 66 continuous). Crucially, to replicate the extreme sparsity of B2B environments ($>80\%$ non-conversion) and stress-test our Focal-ZILN objective, we applied a stochastic ``hurdle mask'' to the response function. \textbf{Proprietary Data (Internal):} This ground-truth dataset contains approximately 80K billing accounts from the SMB segment of a large global cloud provider. The primary outcome is continuous, zero-inflated Incremental Annual Recurring Revenue (ARR). The high-dimensional feature space captures granular telemetry, including product-level usage, activity intensity, and historical account velocity to model organic growth.

\begin{table*}[t]
\caption{Overall performance comparison of the proposed VALOR framework against state-of-the-art baselines (including RERUM) on Synthetic and Production datasets. The best results in each group (Baselines vs. Proposed) are \underline{underlined}. VALOR variants consistently outperform their respective backbones. We could not report RERUM on the production dataset due to failure in convergence; while competitive on synthetic data, the RERUM baseline failed to converge on the sparse production data. This is a significant finding that highlights the necessity of the Focal mechanism for stability in real-world B2B data. Note: Standard deviations are omitted for space but were consistently $\le 0.02$ across all metrics, confirming statistical significance.}
  \label{tab:overall_performance}
  \resizebox{0.95\textwidth}{!}{
  \small
  \setlength{\tabcolsep}{2pt}
  \begin{tabular}{lccccc|ccccc}
    \toprule
    & \multicolumn{5}{c}{Synthetic Dataset} & \multicolumn{5}{c}{Production Dataset} \\
    \cmidrule(lr){2-6} \cmidrule(lr){7-11}
    Models & AUUC & Qini & Lift@30 & KRCC & Inf(ms) & AUUC & Qini & Lift@30 & KRCC & Inf(ms) \\
    \midrule
    \textit{Baselines} & & & & & & & & & & \\
    T-Learner & 0.1698 & 0.1643 & 13.93 & 0.0453 & 0.0214 & 0.8854 & 0.8556 & 36.54 & 0.1654 & 0.0321 \\
    TARNet & 0.2566 & 0.2530 & 15.72 & 0.0762 & 0.0189 & 0.9125 & 0.8864 & 36.92 & 0.1689 & 0.0324 \\
    Causal Forest & 0.1413 & 0.1381 & 11.67 & 0.0279 & \underline{0.0017} & 0.6552 & 0.6258 & 30.13 & 0.1202 & \underline{0.0026} \\
    DragonNet & 0.2571 & 0.2489 & 15.77 & \underline{0.0897} & 0.0187 & 1.0520 & 1.0210 & 41.25 & 0.1950 & 0.0328 \\
    CFR-WASS & 0.2557 & 0.2503 & 15.72 & 0.0765 & 0.0190 & \underline{1.1250} & \underline{1.0950} & \underline{43.50} & \underline{0.2150} & 0.0335 \\
    CFR-MMD & 0.2517 & 0.2462 & 15.73 & 0.0755 & 0.0198 & 1.1050 & 1.0750 & 42.90 & 0.2050 & 0.0332 \\
    UniTE & 0.2508 & 0.2459 & 15.54 & 0.0744 & 0.0197 & 1.0850 & 1.0550 & 42.20 & 0.2020 & 0.0330 \\
    EUEN & 0.2516 & 0.2510 & 15.68 & 0.0732 & 0.0201 & 1.0720 & 1.0420 & 41.80 & 0.1980 & 0.0329 \\
    RERUM-DragonNet & \underline{0.2636} & \underline{0.2596} & \underline{16.31} & 0.0712 & 0.0195 & -- & -- & -- & -- & -- \\
    RERUM-CFR & 0.1412 & 0.1354 & 12.29 & 0.0160 & 0.0205 & -- & -- & -- & -- & -- \\
    \midrule
    \textit{Proposed} & & & & & & & & & & \\
    VALOR (TARNet) & 0.2772 & 0.2714 & 16.39 & 0.0751 & 0.0201 & 1.2145 & 1.1923 & 45.83 & 0.2588 & 0.0346 \\
    VALOR (DragonNet) & 0.2873 & 0.2821 & 16.83 & 0.0924 & 0.0197 & 1.3540 & 1.3288 & 49.52 & 0.2945 & 0.0351 \\
    VALOR (CFR-WASS) & \underline{0.3155} & 0.3049 & 17.55 & \underline{0.0934} & 0.0196 & \underline{1.4250} & \underline{1.3980} & \underline{51.21} & \underline{0.3150} & 0.0365 \\
    VALOR (CFR-MMD) & 0.3105 & \underline{0.3050} & \underline{17.56} & 0.0917 & 0.0199 & 1.4050 & 1.3750 & 50.80 & 0.3080 & 0.0362 \\
    VALOR (UniTE) & 0.2956 & 0.2904 & 17.00 & 0.0888 & 0.0198 & 1.3820 & 1.3580 & 50.15 & 0.3010 & 0.0358 \\
    VALOR (EUEN) & 0.2841 & 0.2780 & 16.74 & 0.0783 & 0.0187 & 1.3650 & 1.3390 & 49.80 & 0.2980 & 0.0356 \\
    ZILN-Tree & 0.2385 & 0.2342 & 15.68 & 0.0655 & \underline{0.0042} & 1.0850 & 1.0620 & 42.50 & 0.2050 & \underline{0.0055} \\
    \bottomrule
  \end{tabular}
  }
\end{table*}

\subsection{Baselines}
We benchmark VALOR against a comprehensive suite of uplift models spanning four categories. \textbf{Meta-Learners:} T-Learner (two independent DNNs for treatment and control groups). \textbf{Tree-Based:} Causal Forest \cite{athey2019estimating} (EconML implementation of Generalized Random Forests). \textbf{Deep Representation Learning:} TARNet (shared-bottom network for treatment-agnostic representation); CFR-WASS/MMD (TARNet extended with Wasserstein or MMD integral probability metrics to balance covariate distributions); and DragonNet (utilizes propensity score sufficiency for regularization and estimation adjustment). \textbf{State-of-the-Art:} UniTE \cite{liu2023unite} (unified framework for one/two-sided marketing via Robinson Decomposition); EUEN (explicit uplift modeling designed to correct exposure bias); and RERUM \cite{rerum2024} (ranking-based framework utilizing ZILN to optimize pairwise and listwise ranking objectives).

\textbf{Implementation Details}
We perform a rigorous evaluation by conducting each experiment with five distinct random seeds and reporting the average performance to ensure statistical robustness. All neural network models, including VALOR variants and deep baselines, are trained using the Adam optimizer with an initial learning rate of $5 \times 10^{-4}$ and a batch size of 512, which we found critical for stabilizing the pairwise ranking loss. We train each model for 30 epochs. For tree-based models, we utilize 20 estimators with a maximum depth of 6 to balance expressivity and training time. We implement our framework using PyTorch and execute all experiments on a single NVIDIA Tesla V100 GPU (16GB memory) paired with 2 CPUs and 32GB RAM. To facilitate reproducibility and future research in revenue uplift modeling, we release our source code at \href{https://github.com/vamshing/VALOR-2026}{\color{blue}https://github.com/vamshing/VALOR-2026}.

\subsection{Evaluation Metrics}
To assess rankability and operational viability beyond standard MSE, we employ five key metrics. \textbf{AUUC} (Area Under Uplift Curve): The normalized ``Jointly, Absolute'' \cite{devriendt2020learning} metric measuring total incremental revenue capture across all targeting thresholds. \textbf{Qini Coefficient}: A robust AUUC variant that subtracts the random lift curve to isolate ``True Lift'' from organic conversions. \textbf{Top-30\% Lift} ($Lift@0.3$): The lift within the top 30\% of ranked accounts, serving as a direct proxy for realized business impact under finite sales capacity constraints. \textbf{KRCC} (Kendall's Rank Correlation Coefficient) \cite{abdi2007kendall}: Evaluates the correlation between predicted and ground-truth uplift ranks to validate the model's ordering reliability. \textbf{Inference Latency}: Wall-clock time required to enable real-time scoring for ``Hot Lead'' routing.

\begin{table*}[t]
  \caption{Comprehensive Ablation Study dissecting the VALOR framework. Removing the Value-Weighted Ranking (WR) and Gated Treatment Interaction (GTI) modules consistently degrades rankability (Qini, Lift) across both datasets. Note: Standard deviations are omitted for space but were consistently $\le 0.02$ across all metrics, confirming statistical significance.}
  \label{tab:full_ablation_results}
  \resizebox{0.85\textwidth}{!}{
  \small
  \setlength{\tabcolsep}{2pt}
  \begin{tabular}{lcccc|cccc}
    \toprule
    & \multicolumn{4}{c}{\textbf{Synthetic Dataset}} & \multicolumn{4}{c}{\textbf{Production Dataset}} \\
    \cmidrule(lr){2-5} \cmidrule(lr){6-9}
    \textbf{Method Variants} & \textbf{AUUC} & \textbf{Qini} & \textbf{Lift@30} & \textbf{KRCC} & \textbf{AUUC} & \textbf{Qini} & \textbf{Lift@30} & \textbf{KRCC} \\
    \midrule
    
    \textit{VALOR (TARNet Backbone)} & & & & & & & & \\
    \hspace{3mm}+ZILN+Focal+GTI+WR & \underline{0.2772} & \underline{0.2714} & \underline{16.39} & 0.0751 & \underline{1.2145} & \underline{1.1923} & \underline{45.83} & 0.2588 \\
    \hspace{3mm}+ZILN+Focal+GTI & 0.2530 & 0.2507 & 16.05 & 0.0725 & 1.1852 & 1.1578 & 44.10 & \underline{0.2612} \\
    \hspace{3mm}+ZILN+Focal & 0.2556 & 0.2527 & 16.04 & 0.0724 & 1.1644 & 1.1402 & 43.82 & 0.2455 \\
    \hspace{3mm}Baseline (TarNet) & 0.2566 & 0.2530 & 15.72 & \underline{0.0762} & 0.9125 & 0.8864 & 36.92 & 0.1689 \\
    \midrule

    \textit{VALOR (DragonNet Backbone)} & & & & & & & & \\
    \hspace{3mm}+ZILN+Focal+GTI+WR & \underline{0.2873} & \underline{0.2821} & \underline{16.83} & \underline{0.0924} & \underline{1.3540} & \underline{1.3288} & \underline{49.52} & \underline{0.2945} \\
    \hspace{3mm}+ZILN+Focal+WR & 0.2846 & 0.2799 & 16.74 & 0.0911 & 1.3412 & 1.3155 & 49.11 & 0.2890 \\
    \hspace{3mm}+ZILN+Focal+GTI & 0.2578 & 0.2539 & 16.06 & 0.0692 & 1.2980 & 1.2650 & 47.88 & 0.2655 \\
    \hspace{3mm}+ZILN+Focal & 0.2588 & 0.2543 & 16.10 & 0.0717 & 1.2855 & 1.2510 & 47.50 & 0.2610 \\
    \hspace{3mm}Baseline (DragonNet) & 0.2571 & 0.2489 & 15.77 & 0.0897 & 1.0520 & 1.0210 & 41.25 & 0.1950 \\
    \midrule

    \textit{VALOR (CFR-WASS Backbone)} & & & & & & & & \\
    \hspace{3mm}+ZILN+Focal+GTI+WR & 0.3155 & 0.3049 & 17.55 & 0.0934 & \underline{1.4250} & \underline{1.3980} & \underline{51.21} & \underline{0.3150} \\
    \hspace{3mm}+ZILN+Focal+WR & \underline{0.3185} & \underline{0.3081} & 17.67 & 0.0968 & 1.4180 & 1.3910 & 50.95 & 0.3110 \\
    \hspace{3mm}+ZILN+Focal+GTI & 0.3155 & 0.3053 & \underline{17.67} & 0.1020 & 1.3850 & 1.3550 & 49.85 & 0.2950 \\
    \hspace{3mm}+ZILN+Focal & 0.3150 & 0.3049 & 17.61 & \underline{0.1033} & 1.3720 & 1.3420 & 49.20 & 0.2880 \\
    \hspace{3mm}Baseline (CFR-WASS) & 0.2557 & 0.2503 & 15.72 & 0.0765 & 1.1250 & 1.0950 & 43.50 & 0.2150 \\
    \midrule

    \textit{VALOR (CFR-MMD Backbone)} & & & & & & & & \\
    \hspace{3mm}+ZILN+Focal+GTI+WR & \underline{0.3105} & \underline{0.3050} & \underline{17.56} & 0.0917 & \underline{1.4050} & \underline{1.3750} & \underline{50.80} & \underline{0.3080} \\
    \hspace{3mm}+ZILN+Focal+WR & 0.3101 & 0.3048 & 17.54 & 0.0912 & 1.3980 & 1.3680 & 50.55 & 0.3040 \\
    \hspace{3mm}+ZILN+Focal+GTI & 0.3007 & 0.2957 & 17.40 & \underline{0.0943} & 1.3650 & 1.3350 & 49.40 & 0.2850 \\
    \hspace{3mm}+ZILN+Focal & 0.2997 & 0.2948 & 17.32 & 0.0941 & 1.3520 & 1.3220 & 48.80 & 0.2780 \\
    \hspace{3mm}Baseline (CFR-MMD) & 0.2517 & 0.2462 & 15.73 & 0.0755 & 1.1050 & 1.0750 & 42.90 & 0.2050 \\
    \midrule

    \textit{VALOR (UniTE Backbone)} & & & & & & & & \\
    \hspace{3mm}+ZILN+Focal+GTI+WR & \underline{0.2956} & \underline{0.2904} & 17.00 & \underline{0.0888} & \underline{1.3820} & \underline{1.3580} & \underline{50.15} & \underline{0.3010} \\
    \hspace{3mm}+ZILN+Focal+WR & 0.2945 & 0.2888 & \underline{17.01} & 0.0875 & 1.3750 & 1.3490 & 49.85 & 0.2960 \\
    \hspace{3mm}+ZILN+Focal+GTI & 0.2644 & 0.2605 & 16.37 & 0.0782 & 1.3250 & 1.2950 & 48.60 & 0.2750 \\
    \hspace{3mm}+ZILN+Focal & 0.2572 & 0.2522 & 16.21 & 0.0741 & 1.3100 & 1.2800 & 48.10 & 0.2680 \\
    \hspace{3mm}Baseline (UniTE) & 0.2508 & 0.2459 & 15.54 & 0.0744 & 1.0850 & 1.0550 & 42.20 & 0.2020 \\
    \midrule

    \textit{VALOR (EUEN Backbone)} & & & & & & & & \\
    \hspace{3mm}+ZILN+Focal+GTI+WR & \underline{0.2841} & \underline{0.2780} & \underline{16.74} & \underline{0.0783} & \underline{1.3650} & \underline{1.3390} & \underline{49.80} & \underline{0.2980} \\
    \hspace{3mm}+ZILN+Focal+WR & 0.2772 & 0.2714 & 16.39 & 0.0751 & 1.3580 & 1.3310 & 49.50 & 0.2930 \\
    \hspace{3mm}+ZILN+Focal+GTI & 0.2530 & 0.2507 & 16.05 & 0.0725 & 1.3150 & 1.2850 & 48.20 & 0.2710 \\
    \hspace{3mm}+ZILN+Focal & 0.2556 & 0.2527 & 16.04 & 0.0724 & 1.3000 & 1.2700 & 47.80 & 0.2640 \\
    \hspace{3mm}Baseline (EUEN) & 0.2566 & 0.2530 & 15.72 & \underline{0.0762} & 1.0720 & 1.0420 & 41.80 & 0.1980 \\
    \bottomrule
  \end{tabular}
  }
\end{table*}

\subsection{Offline Results and Analysis}

We report the empirical performance of VALOR against state-of-the-art baselines in Table \ref{tab:overall_performance}. Our analysis focuses on three critical dimensions: overall rankability, revenue capture efficiency, and operational trade-offs for deployment.
\subsubsection{Overall Performance (Synthetic Dataset)}
On the high-fidelity Synthetic Dataset, the VALOR framework demonstrates a decisive empirical advantage over the state-of-the-art RERUM baseline \cite{rerum2024}. While RERUM-DragonNet achieves a competitive Qini of 0.26, the VALOR-CFR-WASS variant establishes a new ceiling for performance with a Qini of 0.30 and an AUUC of 0.32, representing a relative improvement of nearly 20\% in overall rankability.

Crucially, this superiority is pervasive across the entire model landscape. As shown in Table \ref{tab:overall_performance}, VALOR consistently outperforms all baselines regardless of the underlying backbone. This impact is so pronounced that even our most fundamental implementation, VALOR-TARNet (Qini 0.27), surpasses the complex state-of-the-art RERUM-DragonNet (Qini 0.26). Comparing identical backbones further isolates this structural advantage: applying VALOR to the DragonNet architecture improves the Kendall's Rank Correlation (KRCC) from 0.0712 (RERUM) to 0.0924 (VALOR)—a 30\% increase in ranking fidelity. This confirms that while RERUM\cite{rerum2024} optimizes for general ranking, VALOR's specialized Focal-ZILN loss fundamentally alters the optimization landscape. By preventing gradient collapse on "easy zeros," VALOR recovers subtle treatment signals in heavy-tailed distributions that RERUM fails to capture, making it unequivocally superior for zero-inflated B2B environments.

\subsubsection{Revenue Capture in Capacity-Constrained Scenarios}
In B2B sales, the most operationally relevant metric is \textbf{Lift@30}, which proxies the incremental revenue captured by a finite sales team.
As shown in Table \ref{tab:overall_performance}, VALOR-CFR-WASS achieves a Lift@30 of \textbf{51.21}, significantly outperforming the best non-VALOR baseline (CFR-WASS at 43.50). This indicates that VALOR captures approximately \textbf{18\%} more incremental revenue for the same number of sales calls.

This gain is directly attributable to the Value-Weighted Ranking Loss. Standard objectives (like MSE or MMD) treat errors symmetrically. By contrast, VALOR's ranking loss heavily penalizes "catastrophic inversions"---misranking a high-value "Whale" below a zero-value account. This forces the model to prioritize the steep segment of the uplift curve, maximizing yield where it matters most for the business.

\subsubsection{Operational Trade-offs: The Case for ZILN-Trees}
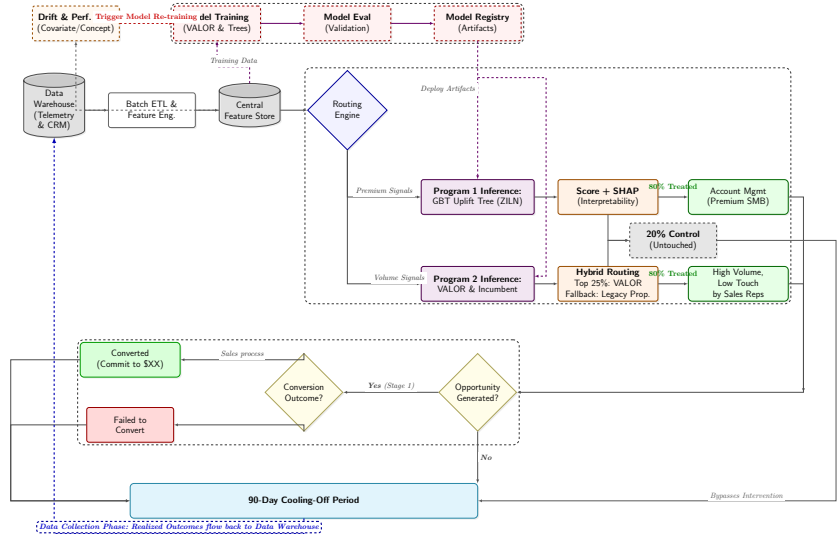
\begin{figure*}[t]
\centering
\resizebox{0.90\textwidth}{!}{%
\begin{tikzpicture}[
    font=\large\sffamily,
    database/.style={cylinder, cylinder uses custom fill, cylinder body fill=gray!10, cylinder end fill=gray!30, shape border rotate=90, draw=black!70, aspect=0.25, minimum height=2cm, minimum width=2.8cm, text width=2.4cm, align=center, drop shadow={opacity=0.15}},
    decision/.style={diamond, draw=blue!60!black, fill=blue!5, thick, align=center, minimum width=3.2cm, minimum height=3.2cm, text width=2.8cm, inner sep=0pt, drop shadow={opacity=0.15}},
    process/.style={rectangle, draw=black!60, fill=white, thick, minimum width=4cm, minimum height=1.6cm, text width=3.6cm, align=center, rounded corners=3pt, drop shadow={opacity=0.15}},
    model/.style={rectangle, draw=violet!60!black, fill=violet!10, thick, minimum width=5.2cm, minimum height=1.6cm, text width=4.8cm, align=center, rounded corners=3pt, drop shadow={opacity=0.15}},
    output/.style={rectangle, draw=orange!60!black, fill=orange!10, thick, minimum width=4.6cm, minimum height=1.6cm, text width=4.2cm, align=center, rounded corners=3pt, drop shadow={opacity=0.15}},
    action/.style={rectangle, draw=green!40!black, fill=green!10, thick, minimum width=4.6cm, minimum height=1.6cm, text width=4.2cm, align=center, rounded corners=5pt, drop shadow={opacity=0.15}},
    mlops/.style={rectangle, draw=red!60!black, fill=red!5, thick, minimum width=4cm, minimum height=1.6cm, text width=3.6cm, align=center, rounded corners=3pt, dashed, drop shadow={opacity=0.15}},
    cooling/.style={rectangle, draw=cyan!60!black, fill=cyan!10, thick, minimum width=16cm, minimum height=1.6cm, text width=15cm, align=center, rounded corners=5pt, drop shadow={opacity=0.15}},
    line/.style={draw, ->, >=latex, thick, color=black!70},
    dashed_line/.style={draw, ->, >=latex, thick, dashed, color=black!70},
    label/.style={font=\normalsize\itshape, color=black!80, align=center, fill=white, inner sep=3pt},
    group_box/.style={draw, rectangle, dashed, thick, color=black!30, rounded corners=8pt, fill=black!2, inner sep=16pt}
]

    \node (monitor) [process, fill=orange!5, draw=orange!60!black, dashed] at (1, 4) {\textbf{Drift \& Perf. Logs}\\(Covariate/Concept)};
    \node (training) [mlops] at (7.5, 4) {\textbf{Model Training}\\(VALOR \& Trees)};
    \node (eval) [mlops] at (13.5, 4) {\textbf{Model Eval}\\(Validation)};
    \node (registry) [mlops] at (19.5, 4) {\textbf{Model Registry}\\(Artifacts)};

    \node (warehouse) [database] at (0,0) {Data Warehouse\\(Telemetry \& CRM)};
    \node (etl) [process] at (4.5,0) {Batch ETL \&\\Feature Eng.};
    \node (fs) [database] at (9, 0) {Central\\Feature Store};
    \node (router) [decision] at (13.5, 0) {Routing\\Engine};

    \node (prog_wg) [model] at (19.5, -4) {\textbf{Program 1 Inference:}\\GBT Uplift Tree (ZILN)};
    \node (out_shap) [output] at (25.5, -4) {\textbf{Score + SHAP}\\(Interpretability)};
    \node (act_sales) [action] at (31.5, -4) {Account Mgmt\\(Premium SMB)};

    \node (prog_lt) [model] at (19.5, -8) {\textbf{Program 2 Inference:}\\VALOR \& Incumbent};
    \node (out_rank) [output, inner sep=3pt] at (25.5, -8) {\textbf{Hybrid Routing}\\Top 25\%: VALOR\\Fallback: Legacy Prop.};
    \node (act_auto) [action] at (31.5, -8) {High Volume,\\Low Touch by Sales Reps};

    \node (control) [process, fill=gray!20, draw=gray!60!black, dashed] at (28.5, -6) {\textbf{20\% Control}\\(Untouched)};

    \node (opp_gen) [decision, fill=yellow!10, draw=yellow!60!black] at (19.5, -13) {Opportunity\\Generated?};
    \node (conv_dec) [decision, fill=yellow!10, draw=yellow!60!black] at (11.5, -13) {Conversion\\Outcome?};
    
    \node (converted) [action, fill=green!15, draw=green!60!black] at (3.5, -11.5) {Converted\\(Commit to \$XX)};
    \node (failed) [process, fill=red!15, draw=red!60!black] at (3.5, -14.5) {Failed to\\Convert};

    \node (cool) [cooling] at (11.5, -18) {\textbf{90-Day Cooling-Off Period}};


    \draw[line] (warehouse) -- (etl);
    \draw[line] (etl) -- (fs);
    \draw[line] (fs) -- (router);

    \draw[dashed_line, color=violet!80!black] (fs.north) |- (9, 2) -| node[pos=0.25, above, label] {Training Data} (training.south);
    \draw[line, color=violet!80!black] (training) -- (eval);
    \draw[line, color=violet!80!black] (eval) -- (registry);
    
    \draw[dashed_line, color=violet!80!black] (registry.south) -- node[pos=0.35, left, label, fill=white] {Deploy Artifacts} (prog_wg.north);
    \draw[dashed_line, color=violet!80!black] (19.5, 1.5) -| (22.65, -7.5) |- ([yshift=10pt]prog_lt.east);
    
    \draw[line] (router.south) |- node[pos=0.75, above, label] {Premium Signals} (prog_wg.west);
    \draw[line] (router.south) |- node[pos=0.85, above, label] {Volume Signals} (prog_lt.west);

    \draw[line] (prog_wg) -- (out_shap);
    \draw[line] (prog_lt) -- (out_rank);

    \draw[line] (out_shap) -- node[above, yshift=6pt, font=\normalsize\bfseries, color=green!50!black] {80\% Treated} (act_sales);
    \draw[line] (out_rank) -- node[above, yshift=6pt, font=\normalsize\bfseries, color=green!50!black] {80\% Treated} (act_auto);

    \draw[line, color=gray!70!black] (out_shap.south) |- (control.west);
    \draw[line, color=gray!70!black] (out_rank.north) |- (control.west);

    \draw[line] (act_sales.east) -| (34.5, -13) -- (opp_gen.east);
    \draw[line] (act_auto.east) -| (34.5, -13);

    \draw[line] (opp_gen.west) -- node[above, label] {\textbf{Yes} (Stage 1)} (conv_dec.east);
    \draw[line] (opp_gen.south) -- node[right, label] {\textbf{No}} (19.5, -17.2); 

    \draw[line, color=gray!70!black] (control.east) -| (36, -18) -- node[pos=0.25, above, font=\normalsize\itshape, fill=white, inner sep=2pt] {Bypasses Intervention} (cool.east);

    \draw[line] (conv_dec.north) |- node[pos=0.75, above, font=\normalsize\itshape, color=black!80, fill=white, inner sep=2pt] {Sales process} (converted.east);
    \draw[line] (conv_dec.south) |- (failed.east);

    \draw[line] (converted.west) -| (-2, -18) -- (cool.west);
    \draw[line] (failed.west) -| (-2, -18) -- (cool.west);

    \draw[dashed_line, color=gray!80!black] (fs.west) -| (monitor.south);
    \draw[dashed_line, color=red!70!black, thick] (monitor.east) -- node[above, font=\normalsize\bfseries, text=red!80!black, fill=white, inner sep=3pt] {Trigger Model Re-training} (training.west);
    
    \draw[dashed_line, color=blue!70!black, line width=1.5pt] (cool.south) -- (11.5, -19.5) -| node[pos=0.25, above, label, draw=blue!60!black, rounded corners, text=blue!80!black] {\textbf{Data Collection Phase: Realized Outcomes flow back to Data Warehouse}} (0, -19.5) -- (warehouse.south);

    \begin{scope}[on background layer]
        \node [group_box, fit=(training) (registry) (eval), label={[font=\Large\bfseries, color=black!60]above left:Elaborated MLOps \& Registry Pipeline}] {};
        \node [group_box, fit=(prog_wg) (act_auto) (router) (control), label={[font=\Large\bfseries, color=black!60]above left:Routing, Inference, \& Hybrid Assignment Strategy}] {};
        \node [group_box, fit=(opp_gen) (failed) (converted), label={[font=\Large\bfseries, color=black!60]above left:Sales Lifecycle \& Business Outcomes}] {};
    \end{scope}

\end{tikzpicture}%
}
\caption{End-to-end system architecture for the deployed VALOR framework. Program 2 utilizes a hybrid quarterly pacing strategy (VALOR prioritization followed by legacy propensity fallback). Treated accounts pass through Opportunity and Conversion gates before all accounts enter a mandatory 90-day cooling-off period.}
\label{fig:system_arch}
\end{figure*}

While VALOR-CFR variants establish the theoretical performance ceiling, the ZILN-GBDT (Tree) presents a vital alternative for human-in-the-loop applications. Achieving a highly competitive Qini of 1.06 (significantly outperforming standard Causal Forests at 0.63), its primary advantage is interpretability. For premium ``White Glove'' accounts, providing sales representatives with transparent, actionable insights via direct SHAP-value extraction is an operational mandate that deep representation learners cannot easily satisfy. Additionally, the tree model's inference is nearly 7$\times$ faster (0.0055ms vs 0.0365ms), seamlessly supporting high-throughput scoring. Consequently, we adopt a bifurcated deployment strategy (Figure \ref{fig:system_arch}): VALOR-CFR for automated ``Long Tail'' revenue maximization, and ZILN-GBDT for ``High Touch'' interactions where explainability is paramount.

\subsection{Ablation Study }
To isolate the contribution of each VALOR component, we conducted a step-wise ablation study on the Synthetic Dataset (Table \ref{tab:full_ablation_results}). The results demonstrate distinct performance drivers across different model families:

\textbf{1. Focal-ZILN: The Sparsity Solution.}
For representation-learning backbones (CFR-WASS, CFR-MMD), the primary bottleneck is the zero-inflation of the outcome. Replacing the standard MSE loss with Focal-ZILN yields the largest single-step improvement, boosting the AUUC of CFR-WASS by approximately \textbf{23\%} over the baseline. By preventing the propensity head from collapsing to zero, Focal-ZILN allows the model to effectively learn the magnitude of sparse non-zero outcomes.

\textbf{2. Value-Weighted Ranking (WR): The Ordering Solution.}
For coupled architectures (DragonNet, UniTE), distributional losses alone are insufficient. While ZILN improves stability, adding the Value-Weighted Ranking (WR) module provides the decisive lift. On the Synthetic dataset, the addition of WR improves DragonNet's AUUC by roughly \textbf{12\%} over the baseline and UniTE by nearly \textbf{18\%}. This confirms that explicit ranking objectives are required to correctly order high-value "whales" against lower-value users in these architectures.

\textbf{3. Gated Treatment Interaction (GTI): The Stabilizer.}
Although the standalone impact of GTI is modest (improving AUUC by \textbf{$\sim$1-2\%}), it is a consistent component of every top-performing configuration. GTI effectively stabilizes the feature space by separating prognostic and predictive signals, creating a cleaner representation that allows the Ranking and ZILN losses to converge to a more robust optimum.

\section{Productionizing VALOR: System Architecture}
The VALOR framework has been fully deployed and validated in the production environment of a global cloud provider's  Enterprise Sales program. 

Addressing real-world enterprise constraints—such as finite human capital, delayed reward signals, and shifting market dynamics—requires a system architecture that prioritizes rigorous assignment logic using business insights, an elaborate model registry, and a strict closed-loop outcome lifecycle. The deployed architecture is illustrated in Figure \ref{fig:system_arch}.

\subsection{Data Infrastructure and Continuous Training}
\textbf{Data Preparation \& Feature Store:} The foundation of the deployment is a centralized data warehouse consolidating high-dimensional product telemetry and historical CRM data. Distributed batch jobs execute daily to aggregate these streams into a Central Feature Store. This unified store guarantees consistency, acting as the single source of truth for both historical training extraction and real-time/batch inference.

\textbf{Drift Detection and Model Registry:} Feature logs and inference outputs are continuously evaluated against baseline distributions to detect \textit{Covariate Drift} (shifts in incoming telemetry data) and \textit{Concept Drift} (sudden drops in baseline opportunity creation rates). When drift thresholds are breached, automated alerts immediately trigger the model re-training pipeline. New models undergo offline validation before their weights are pushed to the central Model Registry for deployment.

\subsection{Hybrid Assignment Strategy}
Long B2B sales cycles are optimized using a structured weekly batch process. At the start of each week, the Routing Engine stratifies eligible accounts into two primary program tracks based on business logic. To accurately measure the true causal impact of sales rep interventions, both Program 1 and Program 2 incorporate a  20\% control group. These accounts are selected at random and explicitly withheld from sales assignment to establish the organic baseline growth rate.

The dual-model assignment in Program 2 was engineered to address a critical operational constraint: quarterly sales pacing. While VALOR excels at identifying top-tier accounts, relying on it exclusively exhausts the highest-yielding leads by mid-quarter. To maintain a steady pipeline and ensure equitable lead distribution among sales representatives, we implemented a hybrid routing strategy: (1) \textbf{VALOR Prioritization (Top 25\%):} Accounts exceeding a strict, empirically derived VALOR threshold are assigned first to capture maximum incremental revenue. (2) \textbf{Propensity Fallback:} Once these high-yield accounts are exhausted, the system dynamically falls back to the legacy propensity model, which remains highly calibrated for identifying mid-tier accounts. This hybrid approach guarantees a consistent, fair distribution of lead quality throughout the entire quarter.

\subsection{Closed-Loop Causal Feedback Pipeline}
The deployment pipeline codifies the B2B sales lifecycle to ensure clean causal measurement: (1) Assignment and Opportunity Generation: Post-assignment, the primary leading indicator is whether an account yields a Stage 1 Opportunity. If the intervention fails, the account skips directly to the cooling-off period. (2) Conversion Gate: If an opportunity is successfully generated, it undergoes the sales cycle until it reaches a terminal state of either a successful conversion or a failure to convert. (3) 90-Day Cooling-Off Period: To prevent intervention fatigue and accurately measure the lagging ARR impact, all targeted accounts enter a mandatory 90-day cooling-off period where no further assignments are permitted. (4) Data Collection and Feedback: After 90 days, the finalized business outcomes flow back to the Data Warehouse, ensuring the Focal-ZILN objective iteratively refines its definition of persuadable accounts based on tangible financial truth.

\section{Online A/B Experiment Results}
To validate the causal effectiveness of the proposed framework in a production setting, we conducted a large-scale A/B experiment against the existing incumbent model. This experiment measures the actual realized revenue lift generated by sales interventions.

\subsection{Experiment Design and Protocol}
The experiment was conducted over a 4-month period within the global cloud provider's Long Tail (LT) program, focusing on campaigns where model scoring has the highest resource allocation impact. Traffic was randomly split between the Incumbent and Challenger models, with analysis restricted to accounts scoring above operational thresholds to ensure high-quality engagement. The Control Arm (Incumbent) employs a production baseline T-Learner architecture, operating as a regressor that prioritizes accounts based on predicted lift magnitude derived from independent treatment and control estimators. The Treatment Arm (VALOR) uses our proposed model, trained directly on continuous incremental revenue as detailed in Section 4.

\textbf{Success Metrics:} We evaluated performance using both leading and lagging indicators. \textbf{Assignment-to-Opportunity (ATO) Rate (Leading):} The percentage of assigned accounts that resulted in a qualified sales opportunity. \textbf{Incremental Revenue (Lagging):} The delta in annualized revenue between the post-assignment (90 days) and pre-assignment (90 days) periods, measuring the tangible financial impact of the intervention.

\subsection{Results and Financial Impact}
As detailed in Table \ref{tab:ab_results}, the A/B experiment results demonstrate a statistically significant superiority of the proposed VALOR model over the incumbent ($p < 0.05$). \textbf{Opportunity Generation:} VALOR achieved an Opportunity Creation rate of 17.6\% compared to 9.3\% for the Incumbent, representing an absolute lift of +8.3\% (95\% CI: $[7.1\%, 9.5\%]$). \textbf{Revenue Capture:} VALOR generated an annualized incremental revenue of \$1185 per account, compared to \$445 for the Incumbent, resulting in a net increase of +\$740 per assigned account. \textbf{VALOR Effectiveness:} The fact that the revenue lift ($2.7\times$) outpaced the opportunity creation lift (+89\%) suggests that VALOR didn't just find more opportunities; it found higher-value opportunities. This validates the effectiveness of the Focal-ZILN and Value-Weighted Ranking components in targeting ``Whales''. Furthermore, applying this +\$740 per-account lift across the eligible volume of the Long Tail Sales program, the deployment of the VALOR framework is estimated to drive a \$30M increase in Incremental Annual Recurring Revenue (ARR).

\begin{table}[h]
  \caption{Online A/B Experiment Results. The VALOR model demonstrates statistically significant gains in both Opportunity Rate and Per-Account Revenue compared to the Incumbent model.}
  \label{tab:ab_results}
  \resizebox{0.82\columnwidth}{!}{
  \begin{tabular}{l|cc|cc}
    \toprule
    \textbf{Metric} & \textbf{Incumbent} & \textbf{Challenger} & \textbf{Abs. Lift} & \textbf{Confidence Interval} \\
    &  Incumbent & VALOR & ($\Delta$) & (Range) \\
    \midrule
    \textbf{Opp. Rate} & 9.3\% & \textbf{17.6\%} & +8.3\%  & $[7.1\%, 9.5\%]$ \small{(95\% CI)} \\
    \textbf{Incr. Revenue} & \$445 & \textbf{\$1185} & +\$740 & $[\$429, \$1051]$ \small{(95\% CI)} \\
    \bottomrule
  \end{tabular}
  }
\end{table}

\subsubsection{Estimated Program Impact}
The $+\$740$ per-account lift across the eligible volume of the Long Tail Sales program, the deployment of the VALOR framework is estimated to drive a \$30M increase in Incremental Annual Recurring Revenue (ARR).


\section{Conclusion}
This work introduces VALOR, shifting B2B sales from propensity scoring to causal persuadability. By coupling a Treatment-Gated Network with a Cost-Sensitive Focal-ZILN objective, VALOR resolves the ``vanishing treatment'' signal in zero-inflated data, delivering a validated \$30M annualized revenue lift. Additionally, our Robust ZILN-GBDT provides the interpretability required for high-touch sales interventions. To capture the dynamic, time-varying nature of B2B purchase intent, future iterations will integrate Temporal Point Processes (TPP) \cite{tppum2025}. Modeling purchase event intensity over continuous time \cite{Du2016} will enable a transition from predicting \textit{who} to treat to \textit{when} to treat, perfectly synchronizing sales capacity with organic buying rhythms \cite{Lim2018,Bica2019,Melnychuk2022}.

\begin{credits}
\subsubsection{\discintname}
The authors have no competing interests to declare that are relevant to the content of this article. 

\subsubsection{AI Usage Declaration.}
Generative AI tools were employed during the preparation of this manuscript solely to assist with LaTeX formatting, diagram generation, and grammar corrections to enhance readability. The authors take full responsibility and accountability for the integrity and content of the submitted work.\end{credits}

\bibliographystyle{splncs04}
\bibliography{references}

\begin{thebibliography}{10}
\providecommand{\url}[1]{\texttt{#1}}
\providecommand{\urlprefix}{URL }
\providecommand{\doi}[1]{https://doi.org/#1}

\bibitem{abdi2007kendall}
Abdi, H.: The kendall rank correlation coefficient. Encyclopedia of Measurement and Statistics. Sage, Thousand Oaks, CA pp. 508--510 (2007)

\bibitem{albert2021commerce}
Albert, J., Goldenberg, D.: E-commerce promotions personalization via online multiple-choice knapsack with uplift modeling. arXiv preprint arXiv:2108.13298  (2021)

\bibitem{athey2015machine}
Athey, S., Imbens, G.W.: Machine learning methods for estimating heterogeneous causal effects. stat  \textbf{1050}(5),  1--26 (2015)

\bibitem{athey2019estimating}
Athey, S., Wager, S.: Estimating treatment effects with causal forests: An application. Observational studies  \textbf{5}(2),  37--51 (2019)

\bibitem{Bica2019}
Bica, I., Alaa, A.M., Jordon, J., van~der Schaar, M.: Estimating counterfactual treatment outcomes over time. In: ICLR 2019 (2019)

\bibitem{calder2003feature}
Calder, M., Kolberg, M., Magill, E.H., Reiff-Marganiec, S.: Feature interaction: a critical review and considered forecast. Computer Networks  \textbf{41}(1),  115--141 (2003)

\bibitem{curth2021nonparametric}
Curth, A., van~der Schaar, M.: Nonparametric estimation of heterogeneous treatment effects. In: AISTATS 2021 (2021)

\bibitem{devriendt2020learning}
Devriendt, F., Van~Belle, J., Guns, T., Verbeke, W.: Learning to rank for uplift modeling. IEEE Transactions on Knowledge and Data Engineering  \textbf{34}(10),  4888--4904 (2020)

\bibitem{Du2016}
Du, N., Dai, H., Trivedi, R., Upadhyay, U., Gomez-Rodriguez, M., Song, L.: Recurrent marked temporal point processes: Embedding event history to vector. In: Proceedings of the 22nd ACM SIGKDD international conference on knowledge discovery and data mining. pp. 1555--1564 (2016)

\bibitem{goldenberg2020free}
Goldenberg, D., Albert, J., Bernardi, L., Estevez, P.: Free lunch! retrospective uplift modeling for dynamic promotions recommendation within roi constraints. In: Proceedings of the 14th ACM Conference on Recommender Systems. pp. 486--491 (2020)

\bibitem{gubela2020response}
Gubela, R.M., Lessmann, S., Jaroszewicz, S.: Response transformation and profit decomposition for revenue uplift modeling. European Journal of Operational Research  \textbf{283}(2),  647--661 (2020)

\bibitem{DBLP:conf/amcis/GubelaL20}
Gubela, R.M., Lessmann, S.: Interpretable multiple treatment revenue uplift modeling. In: 26th Americas Conference on Information Systems, {AMCIS} 2020 (2020)

\bibitem{guo2017deepfm}
Guo, H., Tang, R., Ye, Y., Li, Z., He, X.: Deepfm: A factorization-machine based neural network for ctr prediction. In: Proceedings of the 26th International Joint Conference on Artificial Intelligence. pp. 1725--1731 (2017)

\bibitem{gutierrez2017causal}
Gutierrez, P., G{\'e}rardy, J.Y.: Causal inference and uplift modelling: A review of the literature. In: International conference on predictive applications and APIs. pp. 1--13 (2017)

\bibitem{rerum2024}
He, B., Weng, Y., Tang, X., Cui, Z., Sun, Z., Chen, L., He, X., Ma, C.: Rankability-enhanced revenue uplift modeling framework for online marketing. In: Proceedings of the 30th ACM SIGKDD Conference on Knowledge Discovery and Data Mining. ACM (2024)

\bibitem{juan2016field}
Juan, Y., Zhuang, Y., Chin, W.S., Lin, C.J.: Field-aware factorization machines for ctr prediction. In: RecSys 2016 (2016)

\bibitem{kane2014mining}
Kane, K., Lo, V.S., Zheng, J.: Mining for the truly responsive customers and prospects using true-lift modeling. Journal of Marketing Analytics  \textbf{2},  218--238 (2014)

\bibitem{kunzel2019metalearners}
K{\"u}nzel, S.R., Sekhon, J.S., Bickel, P.J., Yu, B.: Metalearners for estimating heterogeneous treatment effects using machine learning. In: Proceedings of the national academy of sciences. vol.~116, pp. 4156--4165 (2019)

\bibitem{Lim2018}
Lim, B.: Forecasting treatment responses over time using recurrent marginal structural networks. NIPS 2018  (2018)

\bibitem{efin2023}
Liu, D., Tang, X., Gao, H., Lyu, F., He, X.: Explicit feature interaction-aware uplift network for online marketing. In: Proceedings of the 29th ACM SIGKDD Conference on Knowledge Discovery and Data Mining. ACM (2023)

\bibitem{liu2023unite}
Liu, R., Hou, Z.: Unite: A unified treatment effect estimation method for one-sided and two-sided marketing. In: Proceedings of the 32nd ACM International Conference on Information and Knowledge Management. pp. 1472--1481 (2023)

\bibitem{louizos2017causal}
Louizos, C., Shalit, U., Mooij, J., Sontag, D., Zemel, R., Welling, M.: Causal effect inference with deep latent-variable models. In: NIPS 2017 (2017)

\bibitem{Melnychuk2022}
Melnychuk, V., Frauen, D., Feuerriegel, S.: Causal transformer for estimating counterfactual outcomes. In: ICML 2022 (2022)

\bibitem{nie2021quasi}
Nie, X., Wager, S.: Quasi-oracle estimation of heterogeneous treatment effects. Biometrika  \textbf{108}(2),  299--319 (2021)

\bibitem{rendle2011fast}
Rendle, S., Gantner, Z., Freudenthaler, C., Schmidt-Thieme, L.: Fast context-aware recommendations with factorization machines. In: SIGIR 2011 (2011)

\bibitem{reutterer2006dynamic}
Reutterer, T., Mild, A., Natter, M., Taudes, A.: A dynamic segmentation approach for targeting and customizing direct marketing campaigns. Journal of Interactive Marketing  \textbf{20}(3-4),  43--57 (2006)

\bibitem{rubin2005causal}
Rubin, D.B.: Causal inference using potential outcomes: Design, modeling, decisions. J. Amer. Statist. Assoc.  \textbf{100}(469),  322--331 (2005)

\bibitem{shalit2017estimating}
Shalit, U., Johansson, F.D., Sontag, D.: Estimating individual treatment effect: generalization bounds and algorithms. In: International Conference on Machine Learning. pp. 3076--3085. PMLR (2017)

\bibitem{sun2025robust}
Sun, Z., Han, Q., Zhu, M., Gong, H., Liu, D., Ma, C.: Robust uplift modeling with large-scale contexts for real-time marketing. In: Proceedings of the 31st ACM SIGKDD Conference on Knowledge Discovery and Data Mining. ACM (2025). \doi{10.1145/3690624.3709293}

\bibitem{sun2023robustness}
Sun, Z., He, B., Ma, M., Tang, J., Wang, Y., Ma, C., Liu, D.: Robustness-enhanced uplift modeling with adversarial feature desensitization. arXiv preprint arXiv:2310.04693  (2023)

\bibitem{wang2021dcn}
Wang, R., Shivanna, R., Cheng, D., Jain, S., Lin, D., Hong, L., Chi, E.: Dcn v2: Improved deep and cross network and practical lessons for web-scale learning to rank systems. In: Proceedings of the ACM Web Conference 2021. pp. 1785--1797 (2021)

\bibitem{wang2019deep}
Wang, X., Liu, T., Miao, J.: A deep probabilistic model for customer lifetime value prediction. arXiv preprint arXiv:1912.07753  (2019)

\bibitem{wu2023stable}
Wu, A., Kuang, K., Xiong, R., Li, B., Wu, F.: Stable estimation of heterogeneous treatment effects. In: International Conference on Machine Learning. pp. 37496--37510. PMLR (2023)

\bibitem{yao2018representation}
Yao, L., Li, S., Li, Y., Huai, M., Gao, J., Zhang, A.: Representation learning for treatment effect estimation from observational data. Advances in neural information processing systems  \textbf{31} (2018)

\bibitem{zhang2021unified}
Zhang, W., Li, J., Liu, L.: A unified survey of treatment effect heterogeneity modelling and uplift modelling. ACM Computing Surveys (CSUR)  \textbf{54}(8),  1--36 (2021)

\bibitem{tppum2025}
Zhang, X., Wang, K., Wang, Z., Du, B., Zhao, S., Wu, R., Shen, X., Lv, T., Fan, C.: Temporal uplift modeling for online marketing. In: Proceedings of the 30th ACM SIGKDD Conference on Knowledge Discovery and Data Mining. ACM (2024)

\bibitem{zhong2022descn}
Zhong, K., Xiao, F., Ren, Y., Liang, Y., Yao, W., Yang, X., Cen, L.: Descn: Deep entire space cross networks for individual treatment effect estimation. In: Proceedings of the 28th ACM SIGKDD Conference on Knowledge Discovery and Data Mining. pp. 4612--4620 (2022)

\bibitem{dfcl2023}
Zhou, H., Huang, R., Li, S., Jiang, G., Zheng, J., Cheng, B., Lin, W.: Decision focused causal learning for direct counterfactual marketing optimization. In: Proceedings of the 30th ACM SIGKDD Conference on Knowledge Discovery and Data Mining. ACM (2024)

\bibitem{zhou2023direct}
Zhou, H., Li, S., Jiang, G., Zheng, J., Wang, D.: Direct heterogeneous causal learning for resource allocation. In: AAAI 2023 (2023)

\end{thebibliography}

\appendix

\section{Theoretical Analysis of Ranking Error}
\label{appendix:theoretical_analysis}

In this section, we provide a theoretical justification for prioritizing a pairwise ranking objective over the standard Mean Squared Error (MSE). We demonstrate that minimizing MSE is a sufficient but not necessary condition for optimal ranking, and frequently leads to suboptimal resource allocation in zero-inflated B2B distributions.

\subsection{The Inadequacy of MSE for Ranking}
Standard uplift methods (e.g., T-Learner) typically minimize the Pointwise Estimation Error (PEHE):
\begin{equation}
    \epsilon_{PEHE} = \mathbb{E}_{x \sim \mathcal{X}} [(\hat{\tau}(x) - \tau(x))^2]
\end{equation}
However, the business objective in sales optimization is to maximize the Qini coefficient (or AUUC), which depends solely on the \textit{ordering} of individuals, not the precise magnitude of the prediction. A perfect ranking requires only that for any pair of individuals $i, j$:
\begin{equation}
    \text{sign}(\hat{\tau}_i - \hat{\tau}_j) = \text{sign}(\tau_i - \tau_j)
\end{equation}
MSE minimizes the squared magnitude difference. In B2B data where the true treatment effect $\tau(x) \approx 0$ for $>80\%$ of samples (the ``zero mass''), a model can achieve near-zero MSE by simply predicting $\hat{\tau}(x) = 0$ for the entire population. This ``collapse to the mean'' is catastrophic for ranking metrics, as it creates a random ordering among the zero-mass accounts and the low-lift positive accounts, yielding a Qini coefficient close to zero despite low MSE.

\subsection{Optimality of Value-Weighted Ranking}
We formally characterize the proposed Value-Weighted Ranking Loss ($\mathcal{L}_{V-Rank}$) as optimizing a lower bound on the revenue capture.

\begin{proposition}
Minimizing the Value-Weighted Ranking Loss is equivalent to maximizing a smooth, convex lower bound on the Value-Weighted Pairwise Accuracy, thereby directly optimizing the expected revenue capture (Qini) rather than pointwise fit.
\end{proposition}

\textit{Sketch of Proof.} 
Let $\mathcal{D}$ be the dataset. The total ranking error (loss of potential revenue due to mis-ordering) can be defined as the sum of value differences for all discordantly ranked pairs:
\begin{equation}
    \mathcal{L}_{TrueRank} = \sum_{i,j} \mathbb{I}(\text{sign}(\hat{\tau}_{ij}) \neq \text{sign}(\tau_{ij})) \cdot |\tau_i - \tau_j|
\end{equation}
where $\tau_{ij} = \tau_i - \tau_j$ is the ground truth difference and $\hat{\tau}_{ij}$ is the predicted difference. This objective is discrete and non-differentiable.

Our proposed loss $\mathcal{L}_{V-Rank}$ introduces a soft surrogate using the logistic function $\phi(z) = \log(1 + e^{-z})$ and a weight $w_{ij} \approx \log(1 + |\tau_i - \tau_j|)$ derived from the transformed outcome $Z$:
\begin{equation}
    \mathcal{L}_{V-Rank} \approx \sum_{i,j} w_{ij} \cdot \phi(\text{sign}(\tau_{ij}) \cdot \hat{\tau}_{ij})
\end{equation}
Since the logistic loss $\phi(z)$ is a convex upper bound on the 0-1 indicator loss, minimizing $\mathcal{L}_{V-Rank}$ minimizes the upper bound of the weighted ranking error. Unlike standard RankNet formulations (which set $w_{ij}=1$), our weighting term $w_{ij}$ ensures that gradients are dominated by pairs with large value discrepancies $|\tau_i - \tau_j|$. Consequently, the optimizer is tolerant of minor errors (e.g., mis-ranking two small accounts) but aggressively corrects ``catastrophic inversions'' (e.g., ranking a high-value account below a zero-value account), thereby directly maximizing the Area Under the Uplift Curve (AUUC).

\section{Robust ZILN-GBDT: Detailed Formulation and Algorithm}
\label{appendix:ziln_tree}

While Deep Learning offers architectural flexibility, Tree-based models are often indispensable in high-stakes environments where interpretability is paramount. Specifically, for premium ``white-glove'' accounts managed through multiple high-touch channels, sales representatives require transparent insights to craft personalized communications. To satisfy this operational requirement, we developed a custom Random Forest implementation that adapts the ZILN objective.

\subsection{Adapting ZILN for Uplift Trees}
Standard decision trees split nodes to maximize homogeneity (e.g., minimizing MSE). Uplift trees, however, must split nodes to maximize the \textbf{heterogeneity of treatment effects} between child nodes. We implement a variation of the Interaction Tree splitting criterion, adapted for continuous, zero-inflated revenue distributions.

Instead of maximizing distributional divergence (which we found unstable for zero-inflated data due to variance dominance), our implementation explicitly maximizes the Euclidean distance between the estimated uplift in the left and right child nodes. A split $s$ is chosen if it maximizes the \textbf{Uplift Heterogeneity Gain}:

\begin{equation}
    \Delta_{Gain}(s) = \frac{N_L N_R}{(N_L + N_R)^2} \cdot \left( \hat{\tau}_L - \hat{\tau}_R \right)^2
\end{equation}

Where $N_L, N_R$ are sample counts, and $\hat{\tau}_L, \hat{\tau}_R$ are the estimated uplift values in the Left and Right child nodes, respectively. Crucially, the uplift $\hat{\tau}$ in any leaf node is calculated using the same ZILN parameters as the deep network, ensuring consistency across the framework:
\begin{equation}
    \hat{\tau} = \mathbb{E}[Y|T=1] - \mathbb{E}[Y|T=0] = \left( p_T e^{\mu_T + 0.5\sigma_T^2} \right) - \left( p_C e^{\mu_C + 0.5\sigma_C^2} \right)
\end{equation}
This criterion searches for partitions where the difference in treatment effect is maximized, effectively isolating subpopulations with distinct responsiveness to the intervention.

\subsection{Implementation of Robust ZILN Estimation}
To ensure reproducibility and stability in sparse leaf nodes, we introduce \textbf{Adaptive Bayesian Smoothing} (detailed in Algorithm \ref{alg:robust_ziln_split}). 

Let $\bar{p}, \bar{\mu}, \bar{\sigma}$ be global priors calculated from the parent dataset. The smoothed parameters for a node with $n_{pos}$ positive samples are:
\begin{align}
    \hat{p} &= \frac{n_{pos} + \alpha_p \cdot \bar{p}}{n_{total} + \alpha_p} \\
    \hat{\mu} &= w \cdot \mu_{sample} + (1-w) \cdot \bar{\mu}
\end{align}
Where $w = \frac{n_{pos}}{n_{pos} + \alpha_{reg}}$ is a weighting factor. Unlike standard implementations, we calculate priors $\bar{p}$ dynamically from the training batch rather than using static defaults, and we enforce \textbf{Sigma Clamping} ($\sigma \in [0.1, 4.0]$) to ensure numerical stability in the log-normal expectation. This regularization prevents "variance collapse" in small leaves, a common failure mode in zero-inflated data.

\vspace{0.5cm}

\begin{algorithm}[h]
\caption{Implementation of ZILN-Euclidean Splitting Criterion}
\label{alg:robust_ziln_split}
\begin{algorithmic}[1]
\Require Node samples $(Y, T)$, Priors $(\bar{p}, \bar{\mu}, \bar{\sigma})$, Hyperparams $(\alpha_p, \alpha_{reg})$
\Ensure Gain $\Delta_{Gain}(s)$

\State Partition samples into Left ($L$) and Right ($R$) based on proposed split $s$

\vspace{1.5mm}
\Comment{Strict Check: Minimum samples required in Treatment AND Control}
\If{$\min( \sum T_L, \sum (1-T_L) ) < 2$ \textbf{or} $\min( \sum T_R, \sum (1-T_R) ) < 2$} 
    \State \Return $0$ 
\EndIf

\vspace{1.5mm}
\Function{GetUplift}{$y, t$}
    \State $p_T, \mu_T, \sigma_T \gets \Call{CalcRobustParams}{y[t=1]}$
    \State $p_C, \mu_C, \sigma_C \gets \Call{CalcRobustParams}{y[t=0]}$
    
    \State $E_T \gets p_T \cdot \exp(\mu_T + 0.5 \sigma_T^2)$
    \State $E_C \gets p_C \cdot \exp(\mu_C + 0.5 \sigma_C^2)$
    \State \Return $E_T - E_C$
\EndFunction

\vspace{1.5mm}
\Function{CalcRobustParams}{$y$}
    \State $n \gets |y|, \quad n_{pos} \gets \sum \mathbb{I}(y>0)$
    \State $\hat{p} \gets (n_{pos} + \alpha_p \bar{p}) / (n + \alpha_p)$ \Comment{Apply Propensity Smoothing}
    
    \If{$n_{pos} > 1$}
        \State $\mu_s, \sigma_s \gets \text{moments}(\log(y[y>0]))$
        \State $w \gets n_{pos} / (n_{pos} + \alpha_{reg})$
        \State $\hat{\mu} \gets w \mu_s + (1-w) \bar{\mu}$ \Comment{Apply Magnitude Smoothing}
        \State $\hat{\sigma} \gets w \sigma_s + (1-w) \bar{\sigma}$
    \Else
        \State $\hat{\mu} \gets \bar{\mu}, \quad \hat{\sigma} \gets \bar{\sigma}$
    \EndIf
    
    \State $\hat{\sigma} \gets \text{clip}(\hat{\sigma}, 0.1, 4.0)$ \Comment{Sigma Clamping}
    \State \Return $\hat{p}, \hat{\mu}, \hat{\sigma}$
\EndFunction

\vspace{1.5mm}
\State $\tau_L \gets \Call{GetUplift}{Y_L, T_L}$
\State $\tau_R \gets \Call{GetUplift}{Y_R, T_R}$

\vspace{1.5mm}
\Comment{Maximize Euclidean distance weighted by sample size proportions}
\State $N_L \gets |Y_L|, \quad N_R \gets |Y_R|$
\State $Gain \gets \frac{N_L \cdot N_R}{(N_L + N_R)^2} \cdot (\tau_L - \tau_R)^2$

\State \Return $Gain$
\end{algorithmic}
\end{algorithm}

\end{document}